\documentclass[runningheads]{llncs}
\usepackage[T1]{fontenc}
\usepackage{graphicx,verbatim}

\usepackage{amsmath}
\usepackage{amsfonts}
\usepackage[normalem]{ulem}
\usepackage{wrapfig}

\usepackage{xcolor}

\begin{document}
\title{Risk Estimation of Knee Osteoarthritis Progression via Predictive Multi-task Modelling from Efficient Diffusion Model using X-ray Images}

\author{David Butler, Adrian Hilton, Gustavo Carneiro}  
\authorrunning{Butler et al.}
\institute{Centre for Vision, Speech and Signal Processing, University of Surrey \\
    \email{\{d.butler,a.hilton,g.carneiro\}@surrey.ac.uk}}

\maketitle              
\begin{abstract}
Medical imaging plays a crucial role in assessing knee osteoarthritis (OA) risk by enabling early detection and disease monitoring. 
Recent machine learning methods have improved risk estimation (i.e., predicting the likelihood of disease progression) and predictive modelling (i.e., the forecasting of future outcomes based on current data) using medical images, but clinical adoption remains limited due to their lack of interpretability. 
Existing approaches that generate future images for risk estimation are complex and impractical. 
Additionally, previous methods fail to localize anatomical knee landmarks, limiting interpretability. 
We address these gaps with a new interpretable machine learning method to estimate the risk of knee OA progression via multi-task predictive modelling that classifies future knee OA severity and predicts anatomical knee landmarks from efficiently generated high-quality future images.
Such image generation is achieved by leveraging a diffusion model in a class-conditioned latent space to forecast disease progression, offering a visual representation of how particular health conditions may evolve. 
Applied to the Osteoarthritis Initiative dataset, our approach improves the state-of-the-art (SOTA) by 2\%, achieving an AUC of 0.71 in predicting knee OA progression while offering ~9× faster inference time\footnote{This work was partly supported by the Engineering and Physical Sciences Research Council (EPSRC) through grant
EP/Y018036/1.}.


\end{abstract}

\section{Introduction}

\begin{figure}[t!]
    \centering
    \includegraphics[width=0.22\linewidth]{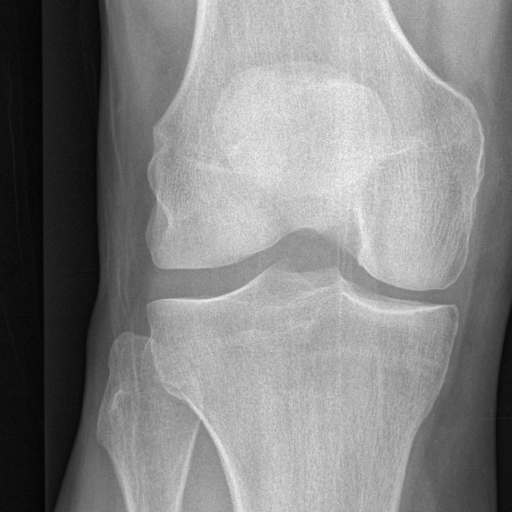}
    \includegraphics[width=0.22\linewidth]{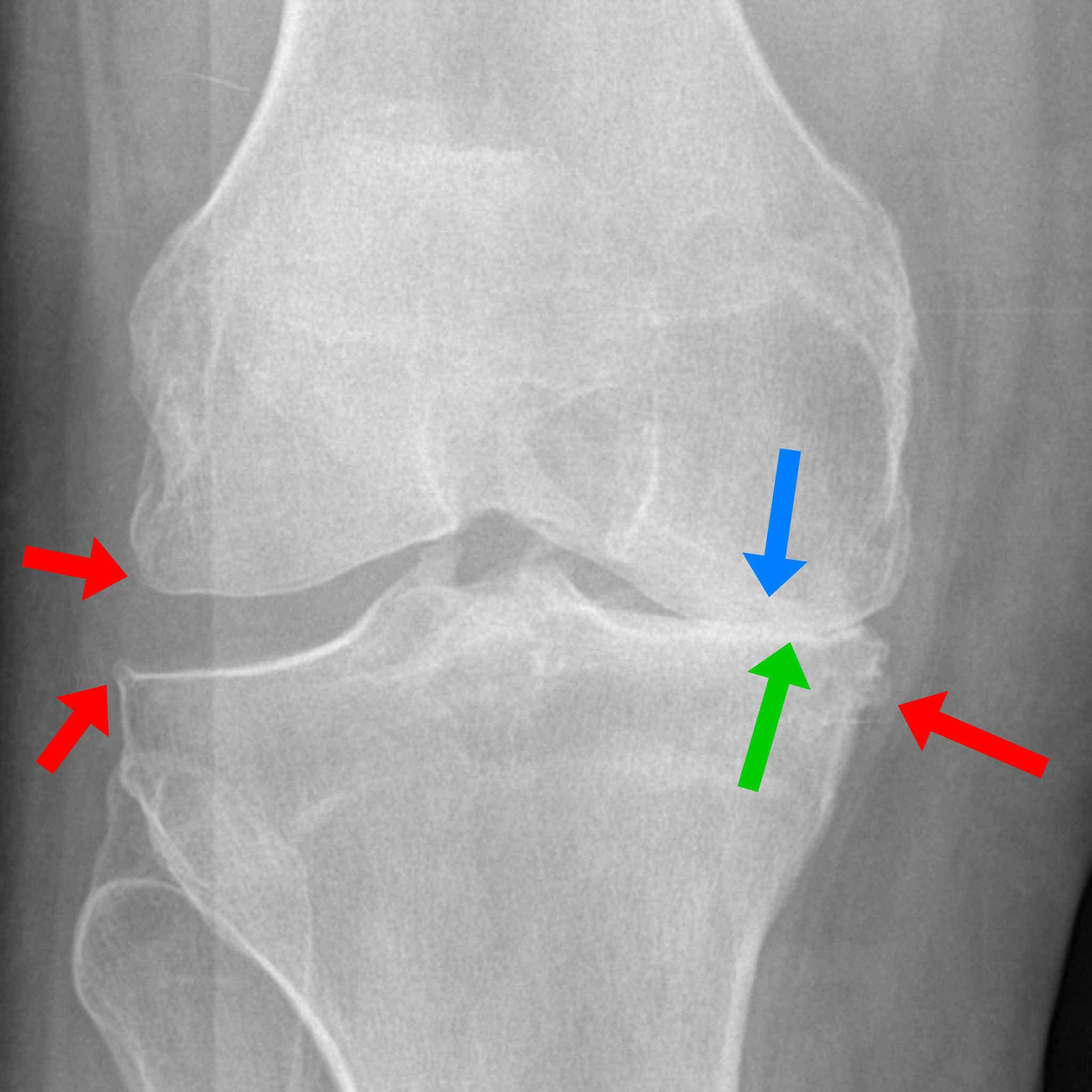}
    \caption{(Left) Example of a 0 KL grade. (Right) Example of a 4 KL grade with osteophytes (red), sclerosis (blue), and bone remodelling (green).}
    \label{fig:kl-examples}
\end{figure}

Knee osteoarthritis (OA) is a degenerative joint disease characterised by cartilage breakdown, bone remodelling, and joint inflammation~\cite{Sinusas2012}. It is a leading cause of disability in older adults, resulting in pain, stiffness, and reduced function. The Kellgren-Lawrence (KL) scale is commonly used to grade osteoarthritis severity, ranging from 0 to 4 based on joint space narrowing, osteophytes, sclerosis, and bone remodelling~\cite{Kohn2016}, as shown in  Fig.~\ref{fig:kl-examples}.  
Early diagnosis enables treatment to alter the disease course~\cite{Chu2012}.

Medical imaging plays a central role in knee osteoarthritis (OA) risk estimation~\cite{Eckstein2012} by analysing tissue changes over time. 
Machine learning techniques compute the likelihood of disease progression~\cite{Tariq2023,Jin2021,Tariq20232,Tariq2022,Cigdem2024}, but most methods generate only numerical scores, offering little visual explanation for clinicians~\cite{Rasheed2022}. 
For instance, if a model predicts OA progression based on X-rays, it is crucial to understand which features, such as OA severity or anatomical landmarks, contribute to this prediction. 
Predictive modelling has been rarely explored, except for~\cite{Han2022}, which employed a highly complex image generation process, limiting clinical practicality and lacking anatomical landmark localization. 
Combining predictive modelling with future image generation and anatomical landmark detection enhances interpretability, fosters trust, and supports informed decision-making.

This paper presents a new interpretable multi-task machine learning method for estimating the risk of knee OA progression by predicting future OA severity grade and anatomical knee landmark localisation from efficiently generated future images.
Such image generation leverages an efficient diffusion model using a class-conditioned latent space to forecast disease progression, offering a visual representation of how such particular health conditions may evolve.
Our key contributions include:
\begin{itemize}
    \item A new interpretable machine learning method for knee OA risk estimation via multi-task prediction modelling for KL classification and anatomical knee landmark localisation using future images generated by a diffusion model; 
    \item A novel, compact, and efficient diffusion model that can generate high quality future OA X-ray images conditioned only by current images.
\end{itemize}
Experiments show that our proposed method has state-of-the-art (SOTA) results on the Osteoarthritis Initiative (OAI) dataset~\cite{Eckstein2012}, a study on knee osteoarthritis, delivering superior risk estimation AUC of $0.71$  while being $\sim9\times$ faster at inference than the previous SOTA\cite{Han2022} that has $0.69$ AUC.

\section{Related Work}

\textbf{Risk Estimation and Predictive Modelling} methods assess risk by predicting clinical events~\cite{Tariq2023,Jin2021,Tariq20232,Tariq2022,Cigdem2024} or forecasting future features~\cite{Lauritzen2023,Nguyen2024,10.1371/journal.pone.0245177,Placido2023,10.1371/journal.pcbi.1006376}. 
While event prediction is useful, it lacks interpretability, as it does not explain underlying causes. 
For instance, multiple plausible progression pathways could lead to mortality, yet these models often do not differentiate between them.  
Similarly, feature prediction models estimate disease onset~\cite{Lauritzen2023,Nguyen2024,10.1371/journal.pone.0245177,Placido2023} or severity~\cite{10.1371/journal.pcbi.1006376}, often using biomarkers~\cite{Placido2023,Nguyen2021} and imaging data~\cite{Nguyen2021,Nguyen2024}. However, their opaque reasoning limits clinical adoption~\cite{Rasheed2022}.

\textbf{Future image synthesis} methods use StyleGAN~\cite{Han2022,Sengupta2020,Alaluf2021}, VAEs~\cite{He2024}, flow-based models~\cite{Liu2024,Shibata2022}, and diffusion models~\cite{10822368}. 
Some rely on an input image and patient information~\cite{Sengupta2020,Alaluf2021,He2024,Liu2024,Shibata2022,Gafuroglu2019,Han2022,Huang2023}, while others omit non-image data like biomarkers~\cite{Liu2024,Shibata2022,Gafuroglu2019,Han2022,Huang2023}. 
Diffusion models now surpass GANs in image quality~\cite{Sinusas2012} but remain computationally demanding and underutilized for disease progression risk estimation~\cite{10822368}. 
In knee OA research, StyleGAN has achieved SOTA accuracy~\cite{Han2022}, yet diffusion models offer superior image quality~\cite{Sinusas2012}. However,~\cite{Han2022} does not generate anatomical knee landmarks, limiting interpretability.

\section{Methodology}

\begin{figure}[t!]
    \centering
    \includegraphics[width=1\linewidth]{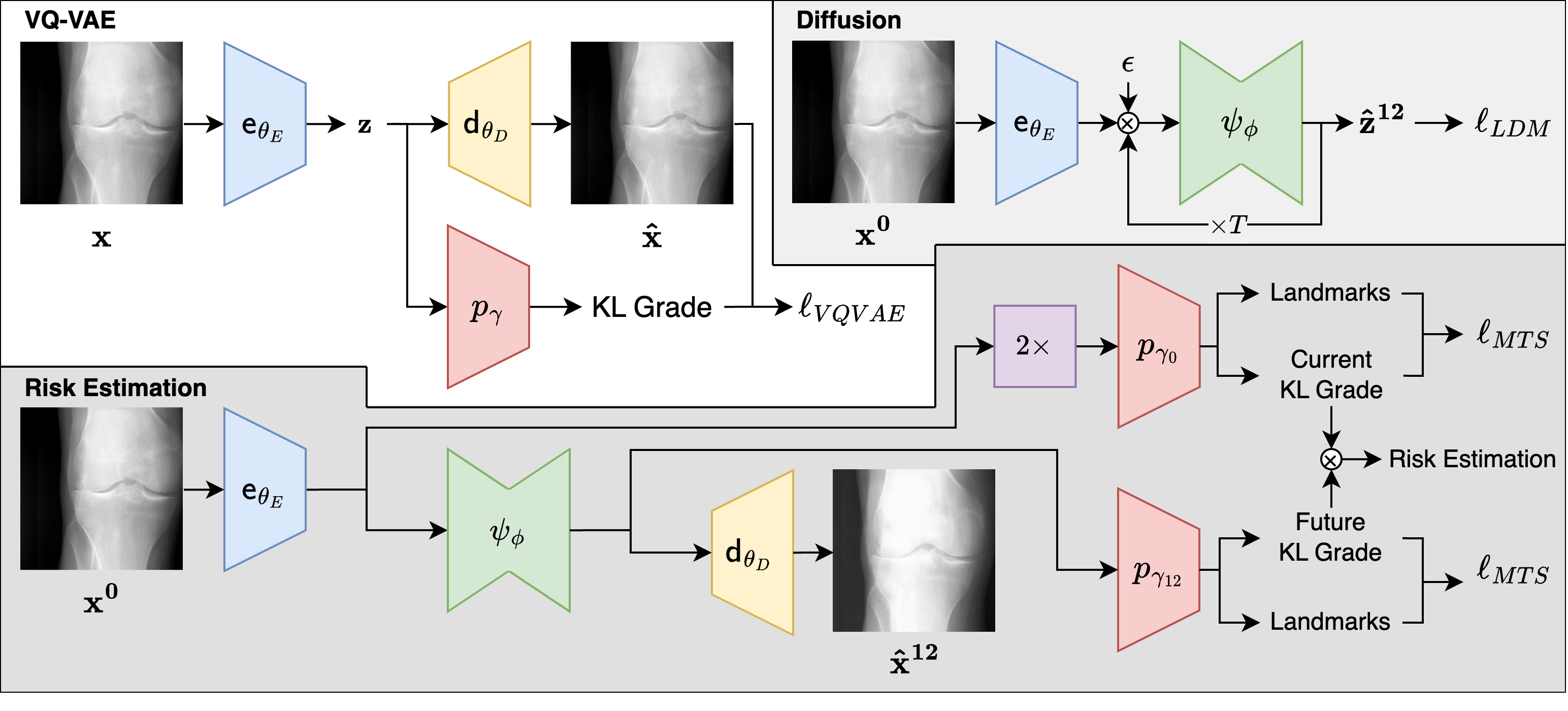}
    \caption{Overview of the method. (Top Left) VQ-VAE training. (Top Right) Diffusion model training. (Bottom) Classifier training \& inference with predicted future image $\mathbf{\hat{x}^{12}}$ and the risk estimated from the KL grades predicted by $p_{\gamma_{0}}$ and $p_{\gamma_{12}}$.}
    \label{fig:method-overview}
\end{figure}

Let $\mathcal{D}= \{ \mathbf{x}_i^0,\mathbf{x}_i^{12},\mathbf{y}_i^0,\mathbf{y}_i^{12}, \{ \mathbf{l}_{i,j} \}_{j=1}^{L} \}_{i=1}^{\left| \mathcal{D} \right|}$  represent the OAI dataset, where $\mathbf{x}^0,\mathbf{x}^{12} \in \mathcal{X} 
\subset \mathbb{R}^{H \times W}$ are knee X-ray images of a patient at an arbitrary point in time, and 12 months afterwards, respectively. 
Corresponding one-hot 5-class KL classifications are \(\mathbf{y}^0,\mathbf{y}^{12} \in \mathcal{Y} \subset \{0,1\}^{5}\). 
The set of \(L\) anatomical knee landmarks at \(\mathbf{x}^{0}\) is \(\{ \mathbf{l}_{i,j} \}_{j=1}^{L} \in \mathcal{L}\), with each landmark \(\mathbf{l}_{i,j} \in \{1,\dots,H\} \times \{1,\dots,W\}\). 
Our model comprises: 1) VQ-VAE for latent image generation, 2) a conditional diffusion model for future latent images, and 3) a multi-task classifier for OA severity prediction and anatomical knee  landmarks localization (Fig.~\ref{fig:method-overview}).

\subsubsection{VQ-VAE:} Future image generation for risk estimation leverages diffusion models, which perform better in latent spaces than in image spaces~\cite{Rombach2022}. 
To generate this latent space, we use VQ-VAE, as it offers superior reconstruction quality and efficiency compared to VQ-GAN~\cite{Esser2021}.
VQ-VAE consists of an encoder \(\mathsf{e}_{\theta_E} : \mathcal{X} \to \mathcal{Z}\) and decoder \(\mathsf{d}_{\theta_D} : \mathcal{Z} \to \mathcal{X}\), with \(\mathcal{Z} \subset \mathbb{R}^Z\) as the latent space, parameterised by \(\theta = \{\theta_E, \theta_D\} \in \Theta\).  
Following~\cite{Rombach2022}, we enhance perceptual quality and classification by integrating a classifier \(p_{\gamma}:\mathcal{Z} \to \Delta^{4}\) for 5-class KL classification, forming a multi-task autoencoder~\cite{Gogna2017}. The model is trained with:  
\begin{equation}
\begin{split}
    \ell_{VQVAE}(\theta,\gamma) =  \mathbb{E}_{\mathbf{x},\mathbf{y}\sim\mathcal{D}} \Big [ & \log(p(\mathbf{x} | \mathbf{z}_q(\mathbf{x}))) + || sg(\mathbf{z}_e(\mathbf{x})) - \mathbf{e} ||_2^2 \\ 
    &+ \beta || \mathbf{z}_e(\mathbf{x}) - sg(\mathbf{e}) ||_2^2 
    - \alpha \sum \mathbf{y}^T \log(p_\gamma(\mathbf{z}_e(\mathbf{x}))) \Big ],
\end{split}
    \label{eq:L_VQVAE}
\end{equation}
where $\mathbf{x}$ is the input image, $\mathbf{z}_e(\mathbf{x}) = \mathsf{e}_{\theta_E}(\mathbf{x})$ is its embedding, $\mathbf{z}_q(\mathbf{x})$ the quantised embedding, $sg(.)$ the stop-gradient operator, $\mathbf{e}$ the nearest codebook entry, $\beta$ controls adherence to the nearest codebook entry, $\alpha$ weights the classification term, $\mathbf{y}$ is the one-hot class label, and $p_\gamma(.)$ the classifier operating in the latent space of the diffusion model. This approach improves the classification accuracy of future synthetic images generated by the diffusion model.

\subsubsection{Conditional Diffusion Model:}
The conditional diffusion model $\mathsf{g}_{\phi} : \mathcal{Z} \to \mathcal{Z}$, parametrised by $\phi \in \Phi$, generates future image embeddings (12 months ahead) conditioned on a patient's current embedding in the latent space $\mathcal{Z}$. 
Following~\cite{Rombach2022}, it learns \(\mathsf{g}_{\phi}(\mathbf{z})\) by iteratively denoising Gaussian noise \(\epsilon \sim N(0, I)\), using a U-Net with \(\mathbf{v}\)-prediction~\cite{salimans2022progressive}, minimising: 
\begin{equation}
    \ell_{LDM}(\phi) = \mathbb{E}_{\boldsymbol{\epsilon},\mathbf{z}^{12},t,\mathbf{z}^0} \left[ || \mathbf{v} - \mathbf{v}_\phi(\mathbf{z}_t^{12}, t, \mathbf{z}^0) ||_2^2 \right],
    \label{eq:L_LDM}
\end{equation}
where $\mathbf{v} = \alpha_t\epsilon - \sigma_t \mathbf{z}^{12}$ is a velocity vector, with $\alpha_t$ and $\sigma_t$ denoting noise and signal proportions at step $t$,  $\mathbf{v}_\phi$ is estimated via U-Net, $\mathbf{z}_t^{12} $ is the latent embedding of the future image, and $ \mathbf{z}^0 $ represents the conditioning image embedding, concatenated with $ \mathbf{z}_t^{12} $ for conditioning. The 
U-Net has four encoding/decoding blocks and a bottleneck, with spatial self-attention in the first three and last three blocks, and channel-wise attention elsewhere. 
Inference model weights are obtained through an exponential moving average during training.

\subsubsection{Risk Estimation via Predictive Modelling:}
Risk estimation uses the conditional diffusion model \(\mathsf{g}_{\phi}(\mathbf{z}^0)\) to generate the future embedding \(\hat{\mathbf{z}}^{12}\) from current image embedding \(\mathbf{z}^0\). 
Two classifiers, denoted by $p_{\gamma_{0}} : \mathcal{Z} \to \Delta^{4}$ and $p_{\gamma_{12}} : \mathcal{Z} \to \Delta^{4}$,
independently classify both $\mathbf{z}^0$ and $\hat{\mathbf{z}}^{12}$. 
The risk, defined as the probability of an increase in KL grade between $\mathbf{z}^0$ and $\hat{\mathbf{z}}^{12}$~\cite{Han2022}, is computed as:  
\begin{align}
    p(y = 1 \mid \mathbf{z}^0, \mathbf{\hat{\mathbf{z}}}^{12}) &= \sum_{c < k} p_{\gamma_{0}}(y^0 = c \mid \mathbf{\mathbf{z}}^0) \cdot p_{\gamma_{12}}(y^{12} = k \mid \mathbf{\hat{z}}^{12}), \label{eq:increase} \\
    p(y = 0 \mid \mathbf{z}^0, \mathbf{\hat{\mathbf{z}}}^{12}) &= \sum_{c \geq k} p_{\gamma_{0}}(y^0 = c \mid \mathbf{z}^0) \cdot p_{\gamma_{12}}(y^{12} = k \mid \mathbf{\hat{\mathbf{z}}}^{12}), \label{eq:stable}
\end{align}
where $y = 1$ indicates an increase in KL grade, $y = 0$ indicates no increase, $y^0$ is the current KL grade, $y^{12}$ is the KL grade after 12 months, and $c, k \in \{0, 1, 2, 3, 4\}$ iterate over KL grades.
The classifier from VQ-VAE multi-task learning serves as an initial model for fine-tuning risk estimation, using
\begin{equation}
    \ell_{CLS}(\gamma_0,\gamma_{12}) = \mathbb{E}_{(\mathbf{x},\mathbf{y}) \sim \mathcal{D}} \left [ - \mathbf{y}^T \log \left(p_{\gamma} \left(\mathbf{y} | \mathbf{z}(\mathbf{x}) \right) \right ) \right ],
    \label{eq:L_CLS}
\end{equation}
where \(\gamma_0\) is estimated from \(\mathbf{x}^{0}, \mathbf{y}^{0}\), and $\gamma_{12}$ from $\mathbf{x}^{12}$ and $\mathbf{y}^{12}$, both in $\mathcal{D}$.
Moreover, $\mathbf{z}^0$ can optionally be upscaled 2 $\times$ with bicubic interpolation at test time, as shown in Fig.~\ref{fig:method-overview} -- we note in the experiments of Sec.~\ref{sec:ablation-study} that such upscaling enables more accurate predictions.

\subsubsection{Multi-task learning}
The multi-task classifier improves classification while predicting anatomical knee landmarks for interpretation. 
It is defined as 
$p_\zeta : \mathcal{Z} \to \Delta^{4} \times \mathcal{L}$, where $\mathcal{L}$ represents $L$ knee landmark coordinates.
Deconvolutional layers are added to the classifier, followed by a 2D SoftArgmax function~\cite{Tiulpin2019}. The model is trained using:  
\begin{equation}
    \ell_{MTS}(\zeta) =  \mathbb{E}_{(\mathbf{x},\mathbf{y},\{\mathbf{l}_j\}_{j=1}^{L}) \sim \mathcal{D}} \left [ - \mathbf{y}^T \log \left[p_{\zeta} \left(\mathbf{y} | \mathbf{z}(\mathbf{x}) \right) \right] + \delta \sum_{j=1}^{L} || \mathbf{l}_j - \mathbf{\hat{l}}_j ||_2^2 \right ],
    \label{eq:L_MTS}
\end{equation}
where $\mathbf{y}$ is the true KL grade for latent image embedding $\mathbf{z}_{e}(\mathbf{x})$, $\mathbf{l}_j = [x_j,y_j]$ is a 2-dimensional landmark coordinate, $\mathbf{\hat{l}}$ is the model’s prediction, and $\delta$ is a weighting hyperparameter.

\subsubsection{Training Algorithm}
Training starts by optimizing VQVAE and its classifier, $p_{\gamma}(.)$ with $\ell_{VQVAE}$ in Eq.~\eqref{eq:L_VQVAE}. The trained VQVAE works as the foundation for training the latent diffusion model, $\mathsf{g}_{\phi}(.)$, with the loss $\ell_{LDM}$ in Eq.~\eqref{eq:L_LDM}. Once trained, the latent diffusion model generates future X-ray images for all dataset samples. Next, classifiers $p_{\gamma_0}(.)$ and $p_{\gamma_{12}}(.)$ are fine-tuned from $p_{\gamma}(.)$ using $\ell_{CLS}$ in Eq.~\eqref{eq:L_CLS}, leveraging ground truth and generated future images, respectively. Alternatively, these classifiers can be optimized with $\ell_{MTS}$ in Eq.~\eqref{eq:L_MTS} to jointly learn KL classification and anatomical knee landmark prediction.

\section{Experiments}

\subsection{Dataset and Assessment}

The Osteoarthritis Initiative (OAI) dataset contains 47,027 knee radiographs from 4,796 patients~\cite{Eckstein2012}, captured at 0-, 12-, 24-, 36-, 48-, 72-, and 96-month intervals. 
Each image is KL-graded, excluding total knee replacements, which cannot be classified. 
Landmark coordinates for $L=16$ joint surface points are provided for 748 images. 
Following~\cite{Tiulpin2019}, all images are cropped to \( 512^2 \) pixels using a landmark prediction model, ensuring full knee visibility. 
Left knee images are flipped for consistency. 
The dataset is split into training (3,772), validation (512), and testing (512) patients. 

Evaluation spans classification, prediction, and risk estimation. Classification involves estimating the current KL class $y^0 \in \{0,1,2,3,4\}$ from $\mathbf{x}^0$ or a latent representation $\mathbf{z}^0$. 
Prediction forecasts KL class \( y^{12} \) 12 months ahead. 
Risk estimation generates a future latent image $\hat{\mathbf{z}}^{12}$ from $\mathbf{z}^0$ using the conditional diffusion model, predicts the KL classifications $y^0$ and $y^{12}$, and calculates the binary probability of KL class progression over 12 months based on Eqs. \ref{eq:increase} and \ref{eq:stable}.

Performance is measured using the mean area under the receiver operating characteristic curve (mAUC), computed as the average of AUC values for each class in a one-vs-rest manner. We compare our method to~\cite{Han2022}, the current SOTA for risk estimation via image generation for knee OA.

\subsection{Training}

The \textbf{VQ-VAE} is trained on the training fold for 5 epochs with a mini-batch size of 8. It uses an Adam optimizer ($\beta_1 = 0.9$, $\beta_2 = 0.999$) and a cosine scheduler (initial LR $10^{-4}$, minimum LR $10^{-6}$). 
Multi-task training with classification uses \(\alpha = 10^{-4}\). The model has a compression ratio of 8, a codebook size of 256, and integrates vector quantization with the decoder.

The \textbf{conditioned diffusion model} is trained on image pairs spaced 12 months apart: \{0,12\}, \{12,24\}, \{24,36\}, and \{36,48\}. Images from 72 and 96 months are excluded due to 24-month gaps. Training runs for 200 epochs with a mini-batch size of 8, using an Adam optimizer $(\beta_1 = 0.9$, $\beta_2 = 0.99$) and a cosine scheduler (initial LR $10^{-4}$, minimum $10^{-6}$). The diffusion process uses 1000 time steps, and sampling applies an exponential moving average of weights with $\gamma = 0.995$.

The \textbf{classifier} is trained on true 0-, 12-, 24-, and 36-month images, and the second classifier is trained on synthetic 12-, 24-, 36-, and 48- month images generated by the diffusion model with 100 time steps for faster inference. Training uses mini-batches of size 8, balanced by whether KL progression occurs. Multi-task classifiers estimates anatomical knee landmarks, trained similarly with a landmark loss weight $\delta = 0.5$.

\subsection{Ablation Study} \label{sec:ablation-study}

\subsubsection{Classification:} 
Tab.~\ref{tab:classification_results} shows lower performance in latent space than image space. However, training the classifier within VQ-VAE mitigates this drop, and fine-tuning further improves results, surpassing image-space classification.


\subsubsection{Prediction:} 
Tab.~\ref{tab:prediction_results} shows lower accuracy than classification (Tab.~\ref{tab:classification_results}) since labels are not directly derived from input images. Latent-space prediction underperforms compared to image space, but training the classifier in VQ-VAE improves results, with fine-tuning further enhancing performance. 
Despite achieving a high mAUC of 0.84, this method predicts only probabilities, making interpretation difficult, and remains less complex than risk estimation, which requires accurate predictions of both \( y^{12} \) and \( y^0 \), as discussed in the next section.


\begin{table}[t]
    \centering
    \begin{minipage}{0.49\linewidth}
        \centering        
        \caption{Ablation study on classification.}
        \label{tab:classification_results}
        \scalebox{0.99}{
        \begin{tabular}{|c|c|}
            \hline        
            Experiment & mAUC \\
            \hline
            \hline
            Image space $p(y^0 | \mathbf{x}^0)$ & 0.82 \\
            \hline
            Latent space $p(y^0 | \mathbf{z}^0)$ & 0.66 \\
            +  VQ-VAE classifier training & 0.70 \\
            + fine-tune VQ-VAE classifier & 0.87 \\
            \hline
        \end{tabular}}
    \end{minipage}%
    \hfill
    \begin{minipage}{0.45\linewidth}
        \centering
        \caption{Ablation study on prediction.}
        \label{tab:prediction_results}
        \scalebox{0.99}{
        \begin{tabular}{|c|c|}
            \hline
            Experiment & mAUC \\
            \hline
            \hline
            Image space $p(y^{12} | \mathbf{x}^0)$ & 0.80 \\
            \hline
            Latent space $p(y^{12} | \mathbf{z}^0)$ & 0.57 \\
            +  VQ-VAE classifier training & 0.71 \\
            +  fine-tune VQ-VAE classifier  & 0.84 \\
            \hline
        \end{tabular}}
    \end{minipage}
\end{table}

\subsubsection{Risk Estimation:} 
Tab.~\ref{tab:prediction_risk_estimation} evaluates risk estimation, \( p(y^{12} > y^0 | \mathbf{z}^0, \hat{\mathbf{z}}^{12}) \). 
The diffusion model generates \(\hat{\mathbf{z}}^{12}\), but image-space evaluation using \(\mathbf{x}^0, \mathbf{x}^{12}\) is also considered for reference. Latent-space performance is lower since the image-space evaluation benefits from the ground truth future images. Training the classifier in VQ-VAE improves results, further enhanced by fine-tuning and multi-task learning with landmark prediction. Upscaling \(\mathbf{z}^0\) at test time significantly boosts performance.

\begin{table}[t]
    \centering
    \caption{Ablation study on risk estimation.}
    \label{tab:prediction_risk_estimation}
    \scalebox{0.99}{
    \begin{tabular}{|c|c|}
        \hline
        Experiment & mAUC \\
        \hline
        \hline
        Image space (ground truth $\mathbf{x}^{12}$) $p(y^{12} > y^0 | \mathbf{x}^0, \mathbf{x}^{12})$ & 0.75 \\
        \hline
        Latent space $p(y^{12} > y^0 | \mathbf{z}^0, \mathbf{\hat{z}}^{12})$ & 0.57 \\
        + VQ-VAE classifier training & 0.60 \\
        + fine-tune VQ-VAE classifier & 0.63 \\
        + multi-task training (classifier+landmark localisation) & 0.65 \\
        + 2 $\times$ upscale $\mathbf{z}^0$  & 0.71 \\
        \hline
    \end{tabular}}
\end{table}

\subsection{Comparison with SOTA}

\begin{figure}[t]
    \centering
    \includegraphics[width=0.24\linewidth]{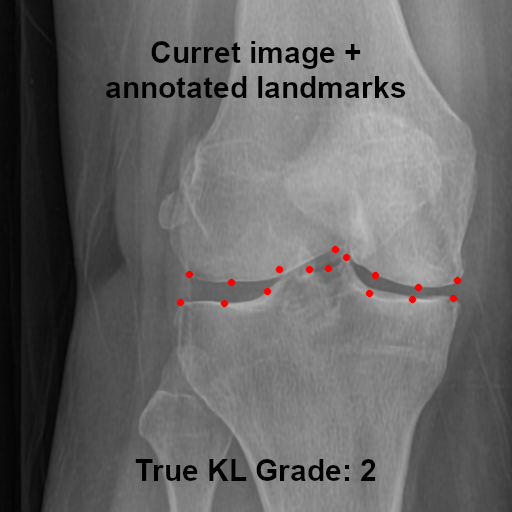}
    \includegraphics[width=0.24\linewidth]{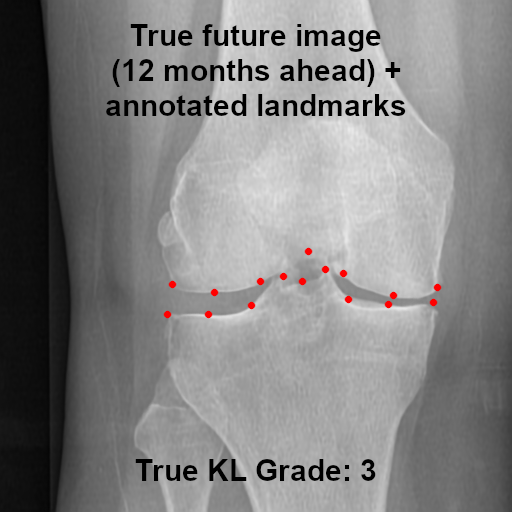}
    \includegraphics[width=0.24\linewidth]{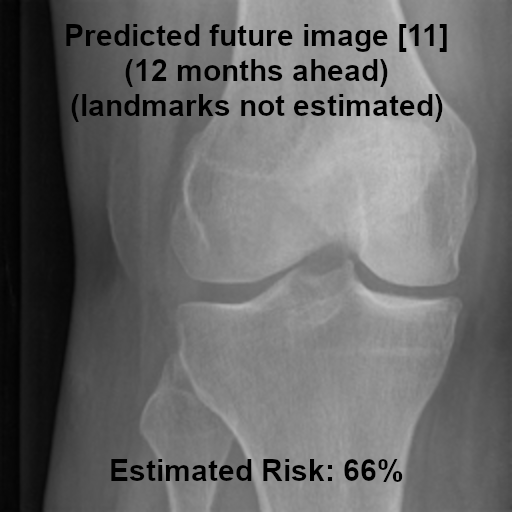}
    \includegraphics[width=0.24\linewidth]{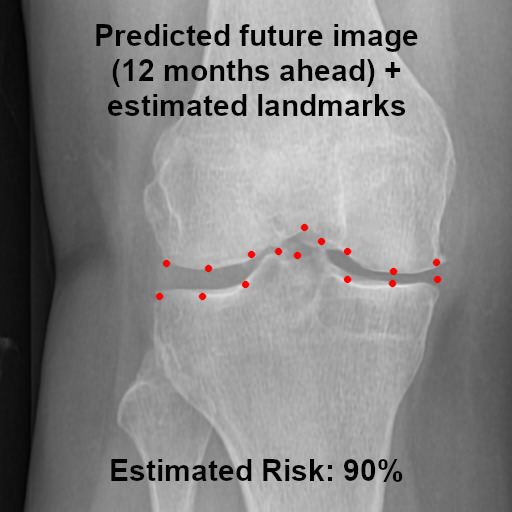}

    \includegraphics[width=0.24\linewidth]{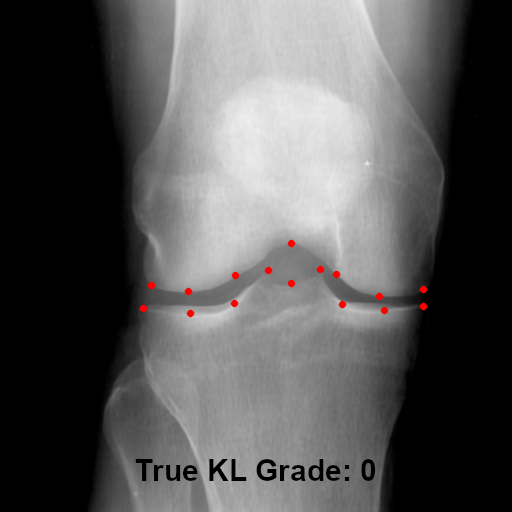}
    \includegraphics[width=0.24\linewidth]{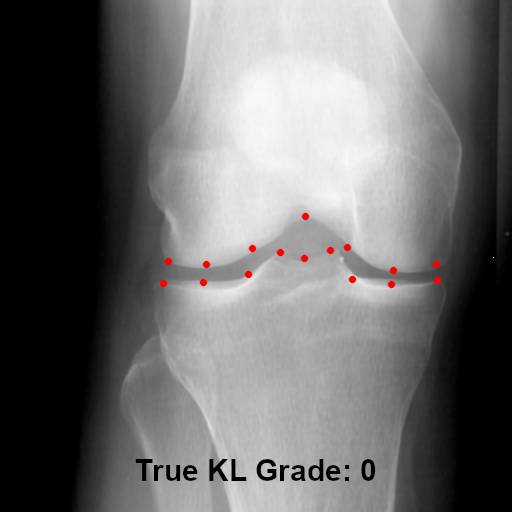}
    \includegraphics[width=0.24\linewidth]{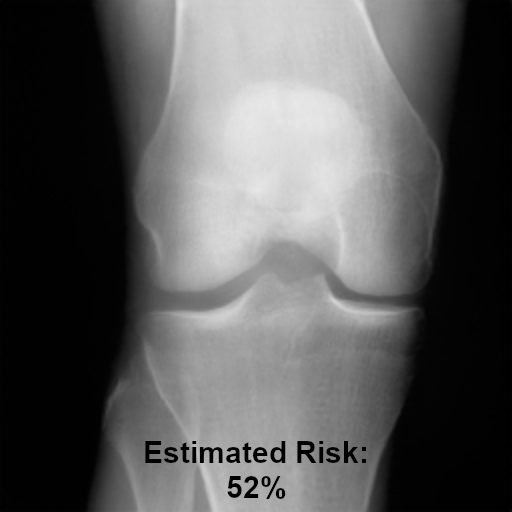}
    \includegraphics[width=0.24\linewidth]{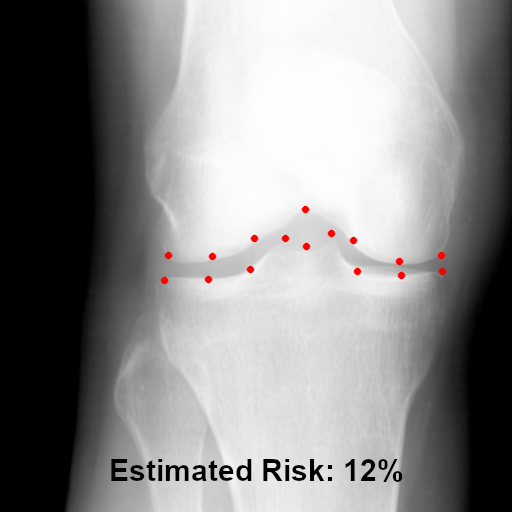}

    \caption{(Left) Current image with annotated landmarks. 
    (Centre Left) True future image (12 months ahead) with annotated landmarks.
    (Centre Right) Predicted future image by~\cite{Han2022} (note that it does not show landmarks)
    (Right) Predicted future image (12 months ahead) with estimated landmarks by our method.
    (Top) Progressing OA with KL grade from 2 to 3. (Bottom) No OA with KL grade 0.
    }
    \label{fig:example_output}
\end{figure}

Our method surpasses SOTA in OAI risk estimation (AUC 0.71 vs. 0.69~\cite{Han2022}) with significantly higher efficiency. 
Our training takes 12.6 hours on a single Nvidia A6000, compared to 114.88 hours on 2× A6000s for~\cite{Han2022}, while our inference is 8.7× faster (2.70s vs. 23.6s per sample). 
Additionally, our approach improves interpretability by not only generating future images but also localizing anatomical knee landmarks, as illustrated in Fig.~\ref{fig:example_output}.  
Beyond generating images that better align with ground truth and providing landmark estimations, our method produces higher-resolution images than~\cite{Han2022}, further enhancing result interpretability.

\section {Discussion \& Conclusion} \label{sec:conclusion}


The proposed method achieves $\sim9\times$ faster inference and a higher risk estimation AUC (0.71 vs. 0.69) than the current SOTA~\cite{Han2022}. By utilising a class-conditioned latent space, our approach enables diffusion models to generate images suitable for predicting future disease progression and allows for a more compact model than the SOTA (our model has 35M vs. 215M parameters in~\cite{Han2022}), and especially in comparison with similar methods utilising diffusion models (35M vs. 1.1B in~\cite{10822368}). Furthermore, incorporating anatomical knee landmarks improves risk estimation while providing additional interpretable outputs.

We find that test-time upscaling of $\mathbf{z}^0$ improves risk estimation for stable low KL scores (0→0,1→1) and increasing high scores (2→3, 3→4) but worsens stable high scores (3→3,4→4) and increasing low scores (0→1, 1→2). We hypothesise this stems from resizing-induced bias, as joint spacing depends on size, whereas osteophytes and sclerosis are more influenced by texture and opacity. Additionally, the model struggles with KL class 1, likely due to its inherent ambiguity—representing doubtful cases rather than mild osteoarthritis—introducing noise that affects neighboring classes (0 and 2). In contrast, clearer symptoms classes (3 and 4) achieve the highest classification accuracy, emphasizing the need for improved label noise handling.

The main limitation of our method is its dependence on class and landmark annotations, which may not always be available. However, landmark annotations are only useful for risk estimation, not for image generation.

For future work, our flexible conditioning mechanism could be extended to multi-image inputs, such as both knees or prior exams, to improve progression modelling. Additionally, recent advancements in conditioning latent diffusion models with non-image data could be explored to enhance predictions. Finally, iterative risk estimation could allow for longer-term forecasting beyond 12 months, improving clinical applicability.


\newpage

\bibliographystyle{unsrt}
\bibliography{library}

\end{document}